\documentclass[letterpaper, 10 pt, conference]{ieeeconf}  

\IEEEoverridecommandlockouts                              

\usepackage{cite}
\usepackage{amsmath,amssymb,amsfonts}
\usepackage{algorithmic}
\usepackage{graphicx}
\usepackage{color,calc}
\usepackage{makecell}
\usepackage{booktabs}  
\usepackage{array}     
\usepackage{ragged2e}  
\usepackage{caption}
\usepackage{multirow}
\usepackage{url}
\usepackage{listings}
\usepackage{xcolor}

\usepackage{subcaption}
\usepackage[ruled,vlined,linesnumbered]{algorithm2e}
\usepackage{amsmath,amssymb}
\usepackage{booktabs}

\lstdefinelanguage{ROSsrv}{
  morekeywords={string,---},
  sensitive=false,
  morecomment=[l]{\#},
}

\lstdefinelanguage{_common}{
  sensitive=false,
  morecomment=[l]{\#},
}

\title{\LARGE \bf
ROSCell: A ROS2-Based Framework for Automated Formation and Orchestration of Multi-Robot Systems
}

\author{Jiangtao Shuai$^{1}$, Marvin Carl May$^{2}$, Sonja Schimmler$^{1, 3}$, Manfred Hauswirth $^{1, 3}$
\thanks{*This research has received funding from the European Union's Horizon RIA research and innovation programme under grant agreement No. 101092908 (SMARTEDGE)}
\thanks{$^{1}$The authors are with the Technical University of Berlin, Germany and $^{3}$Fraunhofer FOKUS, Germany.
        {\tt\small \{firstname.lastname\}@tu-berlin.de}}%
\thanks{$^{2}$Marvin Carl May is with the School of Mechanical \& Aerospace Engineering, Nanyang Technological University, Singapore. {\tt\small marvin.may@ntu.edu.sg} }%
}
\begin{document}
\maketitle
\thispagestyle{empty}
\pagestyle{empty}

\begin{abstract}
Modern manufacturing under High-Mix-Low-Volume requirements increasingly relies on flexible and adaptive matrix production systems, 
which depend on interconnected heterogeneous devices and rapid task reconfiguration. To address these needs, we present ROSCell, 
a ROS2-based framework that enables the flexible formation and management of a computing continuum across various devices. 
ROSCell allows users to package existing robotic software as deployable skills and, with simple requests, assemble isolated cells, automatically deploy skill instances, 
and coordinate their communication to meet task objectives. It provides a scalable and low-overhead foundation for adaptive multi-robot computing in dynamic production environments. 
Experimental results show that, in the idle state, ROSCell substantially reduces CPU, memory, and network overhead compared to K3s-based solutions on edge devices,
highlighting its energy efficiency and cost-effectiveness for large-scale deployment in production settings.
The source code, examples, and documentation will be provided on Github.
\end{abstract}

\section{INTRODUCTION}

The growing demand for flexibility and adaptability in manufacturing under high-mix-low-volume pressure, have motivated matrix production systems 
as a key enabler, granting manufacturers a distinct competitive advantage through efficient resource utilization~\cite{may2021decentralized}. 
The concept of matrix production leverages digitization, IoT software, and flexible machines and robotics to enable cycle-adaptive and decoupled production cells, 
where interconnected equipment and dynamic job scheduling support target-oriented production under changing manufacturing conditions~\cite{chen2023machine, partearroyo2023towards}.
These connected devices generate vast amounts of data which has to be processed and interpreted in a timely and efficient way to meet the increasing autonomy requirements. 
The challenges in this process are: 
A) how to dynamically form independent cells using heterogeneous devices (robots, edge devices, and cloud nodes with diverse specs), to process target-oriented, computationally intensive tasks; 
and B) how to efficiently develop and automatically deploy robotics software while minimizing system downtime.

To address Problem A, the cloud robotics approach~\cite{saha2018comprehensive} leverages the storage space and powerful computational resources of cloud servers to process computation-intensive tasks, 
which, however, faces challenges related to bandwidth limitations and high latency.
To mitigate these issues, the fog robotic approach~\cite{pujol2021fog} introduces an intermediate layer between edge and cloud, 
moving data processing closer to robots to reduce latency and lower the requirements on the cloud network. However, the flexible formation of an independent on-demand computing continuum remains an unsolved issue.

In terms of Problem B, the Robot Operating System (ROS)~\cite{quigley2009ros} provides developers with a comprehensive framework that, through a modular node architecture and a common messaging middleware, 
facilitates rapid reuse of existing packages and tools and thus accelerates the development of robotic applications. 
Its successor, ROS2 improved inter-node communication capabilities via the Data Distribution Service (DDS)-based middleware,
and has seen growing adoption in industrial settings~\cite{he2022ros2, daubaris2024ros2}. 
However, ROS2 by itself does not solve the operational and orchestration complexities associated with large-scale deployments. 
Containerization and microservice patterns, together with orchestration, provide a practicable approach by packaging robotic software into isolated, 
fast-start/stop units that can be managed at scale. Relevant technologies include Kubernetes (K8s) and lightweight distributions such as K3s (designed for resource-constrained or edge environments). 
Nevertheless, the inherent complexity of these platforms remains a practical barrier for robotics developers.

In this paper, we introduce ROSCell to address the above challenges through a combination of ROS2, containerization, and microservice architectures. 
In matrix production, production controllers use ROSCell to assemble devices into internally connected cells, 
while ROSCell can also automatically form size-balanced cells during initialization. 
Developers package robot software as skills; upon a task request, ROSCell orchestrates skill instances onto appropriate devices within a cell and configures the required data pipelines. 
Here, a skill is an atomic operation that implements a specific function (e.g., pose estimation), with multiple variants differing in specifications and deployment options. Details are provided in Sec. ~\ref{sec:roscell}.
The main contributions of this paper are as follows:
\begin{itemize}
\item We present ROSCell, a ROS2-based framework that enables on-demand formation of isolated multi-device cells across the device--edge--cloud continuum and automates containerized robot software deployment.

\item We extend Kubernetes resource allocation to support heterogeneous edge devices, incorporating GPU and storage resources and addressing unified-memory architectures.

\item We evaluate ROSCell against K3s on Raspberry~Pi clusters, 
  demonstrating substantially lower overhead, and demonstrate the applicability of ROSCell in a real robotics application. 
\end{itemize}


\section{RELATED WORK}
\label{sec:sota}
Supporting resource-constrained devices through edge or cloud infrastructures to increase automation is important for advanced manufacturing systems~\cite{yan2017cloud}.
Khan et al.~\cite{khan2025implementation} gathered and processed data from production equipment on cloud servers to enable adaptive scheduling of production processes in real-time.
Xu et al.~\cite{xu2021digital} implemented a framework for digital twin-based cloud robotics to achieve dynamic model updating and manage Robot Control as a Service (RCaaS).
Hussnain et al.~\cite{hussnain2018towards} adopted cloud robotics to achieve intelligent material handling in factories.
These complex scenarios challenge software management in both development and deployment.

To improve scalability for robot software development and deployment, researchers have recently explored the integration of ROS/ROS2 with K8s.
FogROS2~\cite{chen2024fogros2ls} focused on cloud/fog-robotics scenarios and facilitated off-loading individual ROS2 nodes from the edge to cloud computing resources.
However, it requires users to manually configure launch scripts with prior knowledge of instance type, region, and authorization tokens, and its scope is limited to offloading individual ROS2 nodes rather than orchestrating complete task pipelines.
Beyond individual nodes, KubeROS~\cite{zhang2024conquering} deployed entire ROS2-based applications in hybrid computing infrastructures,
enabling a closed-loop workflow with real robots at scale, including benchmark datasets, simulation, real robot validation, and deployment.
However, it focuses on the large-scale deployment of applications rather than cooperation among different robots.
RobotKube~\cite{lampe2023robotkube} proposed event-triggered deployment for collective learning but assumes a pre-configured K8s cluster.
Wen et al.~\cite{wen2024cloud} adopted model-based systems engineering to configure ROS-based applications as microservices, which were then orchestrated in fog robotics systems through K8s.
While ROSCell similarly embraces a microservice-based design philosophy, it eliminates the dependency on Kubernetes entirely, resulting in substantially lower idle-state resource consumption---CPU usage reduced to 2.8\% and memory to 28.4\% of the K3s baseline on worker nodes (see Section~\ref{sec:eva} for details)---making it better suited for energy-sensitive applications.
Furthermore, unlike existing approaches that assume fixed infrastructure topologies and thus cannot support on-demand formation across heterogeneous devices, limiting their applicability to matrix production, ROSCell dynamically forms goal-specific cells in response to human instructions, automatically distributing and executing robot software within isolated cells to achieve flexible resource utilization and improved deployment safety.
The concept of dynamic formation is also central to swarm robotics, where multi-robot behaviors such as pattern formation have been extensively studied~\cite{kaiser2022ros2swarm, dorigo2021swarm}.
However, these approaches generally focus on collective physical behaviors and overlook the computing continuum required for complex robotic applications and industrial-scale cooperation.
ROSCell draws on this principle of dynamic formation but lifts it to the device-edge-cloud continuum: rather than coordinating spatial behaviors, it dynamically assembles heterogeneous computing resources into cells and orchestrates containerized software across them, bridging the gap between swarm-level flexibility and the infrastructure demands of industrial multi-robot systems.

\section{Approach}
\label{sec:roscell}

\begin{figure}[htbp]
  \centering
  \includegraphics[width=0.8\columnwidth]{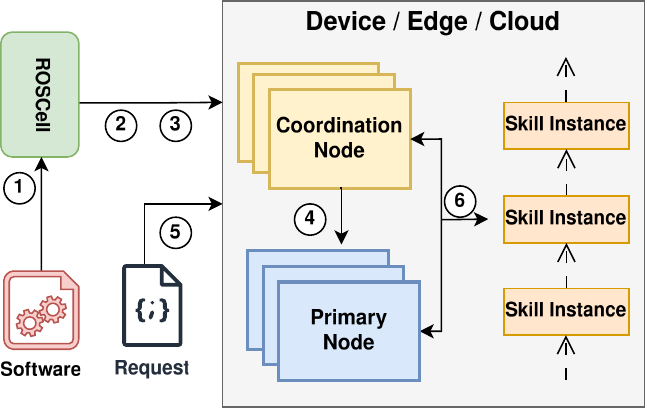}
  \caption{Workflow of ROSCell. 
  Dashed lines show I/O data flows and solid lines information flows. Stacked blocks indicate possible multiple nodes. 
  The gray block illustrates that cell nodes can be any type of resource in the device-edge-cloud continuum.
  The numbered steps show the process of cell formation and software deployment:
  (1) Developers wrap software into ROSCell; 
  (2) The production controller enables devices as ROSCell nodes; 
  (3)\&(4) The coordination node forms its cells via the instruction from the production controller; 
  (5) The operator sends tasks to the coordination node; 
  (6) ROSCell orchestrates and executes the skill instances within the cell.
  }
  \label{fig:runtime-workflow}
\end{figure}

Fig.~\ref{fig:runtime-workflow} shows the workflow of using ROSCell in matrix production: (1) The developer wraps software into ROSCell skills (Sec.~\ref{subsec:skill_orc}); 
(2) The production controller installs the ROSCell on the devices and enables them as ROSCell coordination or primary nodes (Sec.~\ref{subsec:node_role});
(3) The production controller selects cell members and sends the names to a coordination node; 
(4) The coordination node forms its cell (Sec.~\ref{subsec:commu_formation}); 
(5) The operator sends tasks to the coordination node; 
and (6) Within this cell, the coordination node orchestrates the skills to the appropriate nodes, 
where each node deploys the skill instances (Sec.~\ref{subsec:skill_orc} \ref{subsec:skill_manage}).

\subsection{Features and Node Roles}
\label{subsec:node_role}
In the proposed framework, we try to minimize manual configuration requirements. 
In typical environments such as smart factories, numerous devices are already operational and run various deployed software. 
Inspired by \cite{zhang2024conquering}, to manage these devices, 
ROSCell abstracts each as a node with a unique identifier, which is either given by the production controller or automatically generated from its MAC address.
Unlike the traditional production cell, where machines or workspaces are grouped locally, 
we define a cell as a coordinated set of ROSCell nodes to support more flexible application scenarios.
To avoid interfering with existing software configurations, 
any device running a Linux-based OS with Docker installed can be enabled as a ROSCell node using a single command via our launch script, 
with all runtime components deployed in containers.
Since the Docker image is built and cached locally upon the initial launch, subsequent launches reuse the cached image and complete within seconds.

ROSCell defines two node roles: \textit{primary node} and \textit{coordination node}. 
A node's role is determined at startup by passing arguments to the launch script and can be readily reassigned by relaunching with different arguments
(e.g., executing \texttt{./ROSCell\_launch.sh --coordinator} to designate the current device as a coordination node). 
The primary node provides a runtime environment capable of executing skills. 
Each cell contains exactly one coordination node, 
which inherits all primary node capabilities and additionally serves as the cell manager, responsible for handling node join/leave events, 
collecting member status, and dispatching skill instances to appropriate nodes. Details of cell formation are discussed in Sec.~\ref{subsec:commu_formation}.

\subsection{Communication Pattern and Cell Formation}
\label{subsec:commu_formation}
\begin{figure}[htb]
  \centering
  \includegraphics[width=0.9\columnwidth]{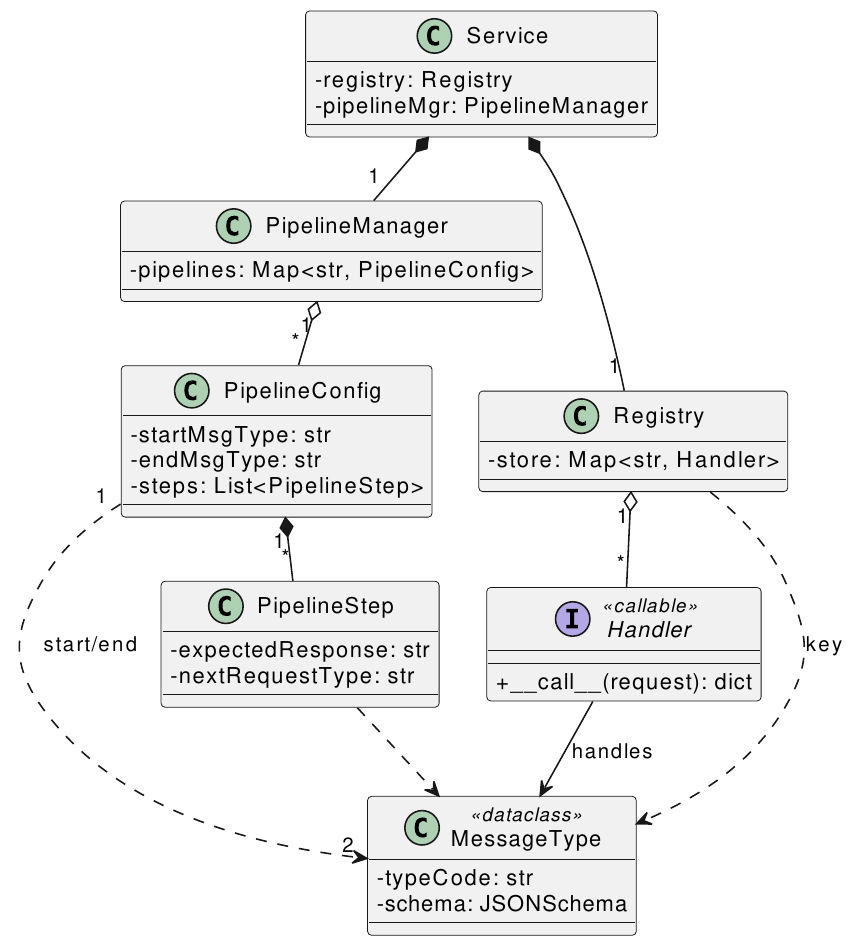}
  \caption{Class diagram of ROSCell message handling mechanism. The circle with C represents a class, while the circle with I represents an interface.}
  \label{fig:communication}
\end{figure}

Once launched, components of a ROSCell node run as ROS2 services inside containers, 
and all interactions, whether intra-node, inter-node, or human-to-node, follow a request-response pattern. 
To handle diverse requests involving complex workflows while keeping the processing logic modular and extensible, 
we define a message-type vocabulary whose models are formally specified using JSON Schema~\cite{pezoa2016foundations}. 
As illustrated in Fig. \ref{fig:communication}, following the registry design pattern, 
each service registers either a dedicated handler for a specific message type, 
or a process pipeline that chains multiple handlers. 
Upon receiving a request, the service first checks whether a registered pipeline matches the incoming message type as its start type; 
if so, the handlers along the pipeline are invoked in order. Otherwise, the service falls back to a matching dedicated handler. 
If no match is found, the request is rejected.

ROSCell leverages the ROS2 DDS discovery mechanism to form cells. 
During startup, users specify a network interface for inter-node communication (e.g., a VPN or LAN address) by passing arguments to the launch script or through interactive prompts. 
For edge-cloud scenarios, ROS2 DDS requires direct IP reachability between participants, which is typically established via VPN when nodes span different networks, as also adopted by FogROS2~\cite{ichnowski2023fogros2} and KubeROS~\cite{zhang2023kuberos}. Since ROSCell itself does not provide VPN services, we assume that such connectivity has been pre-established between devices.
Upon launch, a coordination node automatically configures a DDS discovery server bound to the user-specified address 
and multicasts its presence with a ROS2 service over the network. 
A primary node joins a cell through one of three paths: 
(1) if the coordination node's DDS server address is explicitly passed to the launch script, 
the primary node registers with that coordination node directly; 
(2) if no coordination node is specified, the primary node discovers all available coordination nodes via DDS simple discovery,
queries the size of each cell, and joins the smallest one, enabling size-balanced cell construction during batch initialization---this
multi-cell design is key to scalability: as the number of devices grows, multiple coordination nodes can be deployed to partition the system into
manageable cells, each with a bounded number of members, rather than requiring a single coordinator to manage all nodes;
(3) if no coordination node is currently available, the primary node operates independently and multicasts its identity, 
allowing a coordination node to incorporate it later on demand.

Each primary node runs a built-in HTTP server exposing a REST API dedicated to handling DDS server registration switches. 
Once a primary node joins a cell, its coordination node updates the cell registry, 
including the member list and per-member metadata such as available resources and deployment status. 
Since inter-node communication relies on ROS2 services, 
a primary node registered with a coordination node's DDS server is visible only to nodes within the same cell. 
Across cells, only coordination nodes are mutually visible, and users interact exclusively with the ROS2 services advertised by coordination nodes.

Users can query any coordination node to retrieve cell information and manage node membership dynamically, 
thereby constructing cells that constrain task arrangement within defined scopes, 
such as physical boundaries of production cells or levels in a task hierarchy. 
To transfer a primary node between cells, the user sends a request specifying the primary node and the target coordination node. 
If the request is not sent to the primary node's current coordination node, it is first forwarded there. 
The current coordination node then instructs the primary node via its REST API to switch registration. 
The primary node registers with the target coordination node's DDS server and, upon success, confirms its departure to the original coordination node. 
Both coordination nodes update their cell registries accordingly.

\subsection{Skill Model and Orchestration}
\label{subsec:skill_orc}
To enable automated software deployment across heterogeneous nodes, 
ROSCell introduces a skill model that standardizes the encapsulation and description of executable software.
The skill model consists of two components: a \textit{launcher} and an \textit{engine}.

\begin{figure}[htb]
  \centering
  \includegraphics[width=\columnwidth]{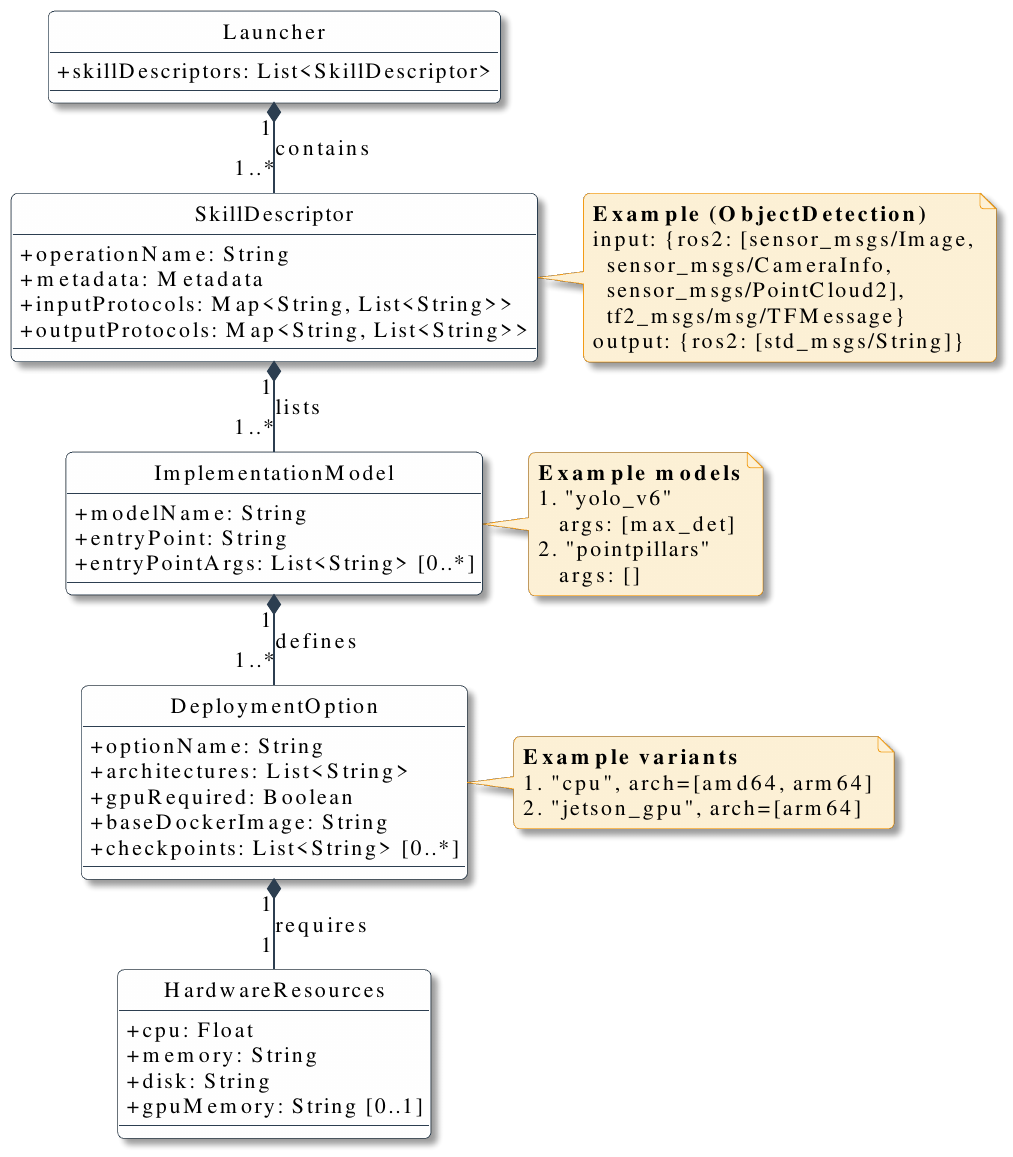}
  \caption{Class diagram of the skill descriptor hierarchy, with an object detection skill as an example. 
  Two implementation models ( \textit{yolo\_v6} and \textit{pointpillars} ) are shown, each offering CPU and GPU deployment options.
  }
  \label{fig:launcher}
\end{figure}

The launcher provides the declarative description of a skill through a machine-readable skill descriptor, 
whose structure is illustrated in Fig.~\ref{fig:launcher}. 
Each skill descriptor specifies the operation name, supported I/O protocols, and one or more implementation models, 
where each model defines its deployment options along with the required hardware resources. 
This structured description allows ROSCell to reason about skill-to-node allocation automatically.

The engine provides the runtime infrastructure that executes a skill. 
Built on the registry pattern, ROSCell currently offers two registered engine types. 
The \textit{PrimitiveBaseEngine} decouples the execution logic into three interchangeable components: 
an input handler that ingests data from a specified source (e.g., ROS2 topic), 
an executor that implements the core processing logic, and an output handler that publishes results. 
Each component is instantiated at runtime based on the protocol and implementation model type declared in the launcher's skill descriptor. 
Developers only need to implement the executor interface, 
consisting of a \texttt{process()} method and a \texttt{release()} method, and register it with the engine. 
The engine then manages the data flow automatically in a continuous input-process-output loop. 
The \textit{ROS2StandaloneEngine} wraps existing ROS 2 applications as-is, 
requiring only declaration in the launcher's skill descriptor without code modification. 
This dual-path design allows ROSCell to accommodate both customized software and ROS2 applications 
as deployable skills with minimal integration effort, providing seamless access to the extensive ROS2 ecosystem 
and enabling rapid reuse of a wide range of existing packages.

With the skill model defined, we now address the orchestration problem, 
i.e., how the coordination node automatically maps a user-submitted task request to concrete skill instances deployed on appropriate nodes. 
Since each skill descriptor declares its resource requirements and compatible architectures, 
this mapping can be formulated as a resource-aware allocation problem.
Inspired by Kubernetes (K8s) scheduling~\cite{carrion2022kubernetes},
ROSCell implements a scheduler on the coordination node
that allocates skill instances across cell nodes while balancing resource utilization,
which is critical for heterogeneous robotic systems where devices vary widely in computational capacity.
The problem is formulated as follows.

Given a user request comprising an ordered execution pipeline $E = (e_1, e_2, \ldots, e_n)$ with $n$ tasks,
where each task specifies a task ID, I/O data types and endpoints, operation name, implementation model name, 
and an optional deployment preference, 
the coordination node allocates suitable skill instances across its cell $V = \{v_1, v_2, \ldots, v_m\}$ with $m$ nodes.
According to the skill descriptor, 
it first associates each task $e_i$ with a set of feasible allocation candidates 
$C_i = \{(d, v) \mid d \in D_i, v \in V, \texttt{compatible}(d, v)\}$, 
where $D_i$ denotes the set of deployment options supported by the implementation model specified in $e_i$ 
(filtered to a singleton if the user explicitly specifies a deployment preference), 
and $\texttt{compatible}(d, v)$ checks whether node $v$ satisfies the resource specs of deployment $d$.
The coordination node then constructs the space of global allocation schemes via the Cartesian product 
$\Phi = C_1 \times C_2 \times \cdots \times C_n$, 
where each element $\mathcal{A} = \bigl((d_1, v'_1), (d_2, v'_2), \ldots, (d_n, v'_n)\bigr) \in \Phi$ 
represents a complete allocation assigning every task $e_i$ to a specific deployment $d_i$ on a specific node $v'_i \in V$.
In practice, the search space is effectively pruned by the compatibility check and optional user-specified deployment preferences,
keeping the enumeration tractable for typical cell sizes (tens of nodes with a bounded number of concurrent tasks).
To select the optimal allocation from $\Phi$, we extend the K8s {\it BalancedResourceAllocation} scoring strategy~\cite{carrion2022kubernetes},
which originally considers only CPU and RAM.
Our extension incorporates disk and GPU memory, as many modern robotic applications, particularly learning-based methods, demand GPU compute and significant storage resources.
The objective minimizes the variance of resource fractions per allocation (to avoid bottlenecks on any single resource type)
and the variance of load across nodes (to prevent hotspots), formulated as:

\begin{equation}\label{eq:compact_obj}
\begin{split}
\mathcal{A}^*=\arg\min_{\mathcal{A}\in\Phi}\Bigg\{
&\underbrace{\frac{1}{n}\sum_{i=1}^{n}\bigg(\frac{1}{|\mathcal{R}_i|}
\sum_{\mathsf{r}\in\mathcal{R}_i}(f_i^{\mathsf{r}}-\bar{f}_i)^2
+ \beta_i\bigg)}_{\bar{\sigma}(\mathcal{A})}\\
&+\;\alpha\underbrace{\frac{1}{|V'|}\sum_{v\in V'}
\Big(\ell_v - \bar{\ell}\Big)^2}_{\sigma_{\mathrm{load}}(\mathcal{A})}
\Bigg\}
\end{split}
\end{equation}

with:

\begin{equation}\label{eq:sigma_sym}
\begin{aligned}
\mathcal{R}_i &= \begin{cases}
\{\mathrm{cpu},\mathrm{mem},\mathrm{disk},\mathrm{gmem}\} & \text{if } d_i \text{ requires GPU,}\\
\{\mathrm{cpu},\mathrm{mem},\mathrm{disk}\} & \text{otherwise,}
\end{cases}\\[6pt]
\bar{f}_i &= \frac{1}{|\mathcal{R}_i|}\sum_{\mathsf{r}\in\mathcal{R}_i} f_i^{\mathsf{r}},\qquad
\beta_i = \begin{cases} -0.01 & \text{if } d_i \text{ requires GPU,}\\ 0 & \text{otherwise,}\end{cases}\\[6pt]
\ell_v &= \tfrac{1}{2}\big(\ell_v^{\mathrm{cpu}}+\ell_v^{\mathrm{mem}}\big), \qquad
\bar{\ell}=\frac{1}{|V'|}\sum_{v\in V'}\ell_v, \\[6pt]
\ell_v^{\mathrm{cpu}} &= \frac{u_v^{\mathrm{cpu}}+\sum_{(e,v,d)\in\mathcal{A}} r_{e,d}^{\mathrm{cpu}}}{c_v^{\mathrm{cpu}}},\\[6pt]
\ell_v^{\mathrm{mem}} &= \frac{u_v^{\mathrm{mem}}+\sum_{(e,v,d)\in\mathcal{A}} r_{e,d}^{\mathrm{mem}}}{c_v^{\mathrm{mem}}}
\end{aligned}
\end{equation}

subject to:
\begin{equation}\label{eq:constraints}
\begin{aligned}
&\forall v\in V,\; \forall\, \mathsf{r}\in \mathcal{R}_i \setminus \{\mathrm{gmem}\}:\quad 
\sum_{\substack{(e,v,d)\in\mathcal{A}}} r_{e,d}^{\mathsf{r}} \le c_v^{\mathsf{r}} - u_v^{\mathsf{r}},\\[6pt]
&\forall v\in V\setminus V_{\mathrm{um}},\; \forall\, g\in G_v:\quad 
\sum_{\substack{(e,v,d)\in\mathcal{A},\\ \mathrm{gpu}(d)=g}} r_{e,d}^{\mathrm{gmem}} \le c_g^{\mathrm{mem}} - u_g^{\mathrm{mem}},\\[6pt]
&\forall v\in V_{\mathrm{um}}:\quad 
\sum_{(e,v,d)\in\mathcal{A}}\big(r_{e,d}^{\mathrm{mem}}+r_{e,d}^{\mathrm{gmem}}\big)
\le c_v^{\mathrm{mem}} - u_v^{\mathrm{mem}}.
\end{aligned}
\end{equation}

where $\mathcal{R}_i \subseteq \{\mathrm{cpu},\mathrm{mem},\mathrm{disk},\mathrm{gmem}\}$ is the set of resource types relevant to allocation $i$, depending on whether deployment $d_i$ requires GPU.
$\beta_i$ is a tunable hyperparameter that introduces a small bonus favoring GPU-accelerated deployments when available, since GPU variants typically offer higher throughput for compute-intensive skills; we set $\beta_i = -0.01$ in our current implementation.
$\alpha$ is a tunable hyperparameter controlling the weight of cross-node load balancing relative to per-allocation resource balance; we set $\alpha = 0.1$ in our experiments.
$V' \subseteq V$ is the subset of nodes that host at least one task under $\mathcal{A}$.
$G_v$ is the set of discrete GPUs on node $v$ (applicable to nodes with dedicated GPU memory).
$V_{\mathrm{um}} \subseteq V$ is the set of nodes with unified memory architecture (e.g., NVIDIA Jetson), where CPU and GPU share a single memory pool.
$c_v^{\mathsf{r}}$ and $u_v^{\mathsf{r}}$ are the total capacity and current usage of resource $\mathsf{r}$ on node $v$, respectively. $c_g^{\mathrm{mem}}$ and $u_g^{\mathrm{mem}}$ are the total and used memory of GPU $g$.
$r_{e,d}^{\mathsf{r}}$ is the amount of resource $\mathsf{r}$ requested by task $e$ under deployment $d$.
For each allocation $i = (e, v, d)$, the resource fraction is:

\begin{equation}\label{eq:resource-fraction}
f_i^{\mathsf{r}} = \frac{r_{e,d}^{\mathsf{r}}}{c_v^{\mathsf{r}} - u_v^{\mathsf{r}} - \displaystyle\sum_{\substack{(e',v,d') \in \mathcal{A},\\ e' \prec e}} r_{e',d'}^{\mathsf{r}}}
\end{equation}

$e' \prec e$ denotes that $e'$ precedes $e$ in the user-specified task ordering within $E$.
For unified-memory nodes ($v \in V_{\mathrm{um}}$) requiring GPU, fractions are computed against the shared memory pool after reserving the counterpart's demand (i.e., $f_i^{\mathrm{mem}}$ uses available memory minus GPU reservation, and vice versa).

\subsection{Skill Deployment and Lifecycle Management}
\label{subsec:skill_manage}

With the allocation $\mathcal{A}^*$ determined, the coordination node proceeds to deploy and manage skill instances across the cell. 
The coordination node dispatches each deployment to the corresponding node (deployments allocated to itself are handled locally). 
Upon receiving a deployment, the node dynamically generates a Dockerfile from the deployment option's base image, the skill engine, 
and the launcher's entry point, and builds a container image automatically. Runtime parameters, including I/O endpoints and checkpoints, 
are injected via environment variables, decoupling task configuration from the container image. Consequently, 
when a node receives a deployment whose image has been previously built, it reuses the cached image and only updates the environment variables, 
significantly reducing startup time.

Each running container constitutes a skill instance identified by a unique instance ID. 
Instance metadata, including the task ID, instance ID, and deployment image, 
is recorded on the node and synchronized with the coordination node. 
Running each skill instance in an isolated container prevents environment conflicts between concurrent instances and enables straightforward lifecycle management. 
The coordination node can query the status of skill instances on each node to maintain a global view of active and inactive deployments. 
Upon receiving a task termination request, the coordination node forwards the request to the relevant nodes by task ID. 
Each node then removes the corresponding skill instances, releases the occupied resources, and synchronizes the updated status with the coordination node.

\section{Evaluation}
Our evaluation focuses on the overhead introduced by the ROSCell framework itself, measured when nodes are in idle state, 
i.e., before/after deployment without active request processing. 
This reflects the typical operating condition in practice, 
where nodes spend the majority of their time maintaining the cell structure and running deployed skill instances 
rather than actively processing new requests.
The resource consumption of individual skill instances is excluded from this evaluation, 
as it is determined by the user's choice of software and task characteristics (e.g., model complexity, input data modality) rather than by the framework. 
By isolating the framework-level overhead, we provide a fair and reproducible baseline independent of application-specific workloads.

\label{sec:eva}
\subsection{System Overhead}
We first evaluate the resource footprint of ROSCell, 
as it is critical for resource-constrained edge devices and energy-efficient manufacturing environments, 
where minimizing system overhead directly translates into improved performance and reduced operational cost.
As discussed in Section~\ref{sec:sota}, most existing frameworks are built upon K8s. 
K3s, a lightweight distribution of K8s designed for resource-constrained environments, can be regarded as the lower bound for such systems. 
Therefore, we directly compare ROSCell with K3s in terms of CPU utilization, memory usage, and disk space.

\begin{table}[htbp]
\centering
\begin{tabular}{clccccc}
\toprule
\multirow{2}{*}{Scenario} & \multirow{2}{*}{Software} 
& \multicolumn{2}{c}{CPU [\%]} 
& \multicolumn{2}{c}{Memory [MiB]} 
& Disk [MB] \\
\cmidrule(lr){3-4} \cmidrule(lr){5-6} \cmidrule(lr){7-7}
& & Pi4 & Pi5 & Pi4 & Pi5 & Pi4/Pi5 \\
\midrule
\multirow{2}{*}{I} 
 & ROSCell & \textbf{0.20} & \textbf{0.07} & \textbf{294} & \textbf{214} & 1027 \\
 & K3s     & 7.1 & 1.4 & 1036  & 743 & \textbf{311}  \\
\cmidrule{3-7}
\multirow{2}{*}{II} 
 & ROSCell & \textbf{0.64} & \textbf{0.18} & \textbf{343} & \textbf{262} & \textbf{1027} \\
 & K3s     & 38.6 & 11.2 & 1788 & 1316 & 1111 \\ 
\bottomrule
\end{tabular}
\caption{\textbf{Comparison of system overhead under equivalent configurations.} Scenario I: a single Raspberry Pi operating as a ROSCell primary node or a K3s worker node. Scenario II: a Raspberry Pi serving as a ROSCell coordination node or a K3s master node, each managing five additional nodes.}
\label{tab:system-overhead}
\end{table}

We evaluated the system overhead using two clusters, each composed of six Raspberry Pi devices (six Pi 4 and six Pi 5), 
and each configured with one master and five workers. Master refers to either the K3s master or the ROSCell coordination node, 
while worker refers to either K3s workers or ROSCell primary nodes. All devices communicated over wired Ethernet. 
We used v1.33.5+k3s1 of K3s.
To reduce device-specific variance, we applied a rotation strategy: after forming a cluster and collecting 120 seconds of resource data, 
we reassigned a different device as master and repeated the measurement, ensuring that each device served once as master with five workers. 
The median values of all metrics are presented in Table~\ref{tab:system-overhead}.

Regarding disk space usage, ROSCell worker nodes consume more storage than K3s workers, whereas ROSCell masters require less storage than K3s masters. 
For CPU and memory consumption—both in the worker role (Scenario I) and the master role (Scenario II)—ROSCell shows significantly lower overhead than K3s. 
Notably, the K3s master node consumes substantially more resources than its workers (e.g., on the Pi 5, 8$\times$ more CPU and 1.77$\times$ more memory), 
and its disk usage grows considerably with cell size, limiting both the scalability of edge cells and the range of devices eligible to serve as master nodes.
In contrast, ROSCell's lower power usage and reduced role dependency in terms of resource usage make it more suitable for dynamic and rapidly changing production environments.

\subsection{Network Throughput}

\begin{figure}[htbp]
    \centering

    \begin{subfigure}{\columnwidth}
        \centering
        \includegraphics[height=4cm]{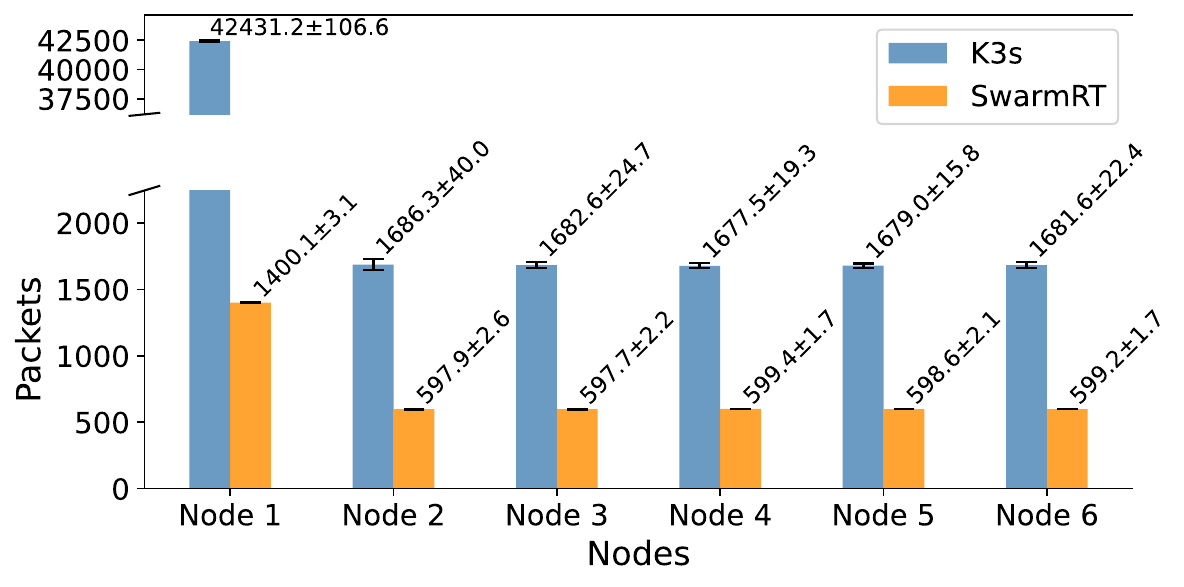}
        \caption{Packet distribution over nodes in 10 min}
        \label{fig:todo_packet}
    \end{subfigure}

    \vspace{0.3em}

    \begin{subfigure}{\columnwidth}
        \centering
        \includegraphics[height=4cm]{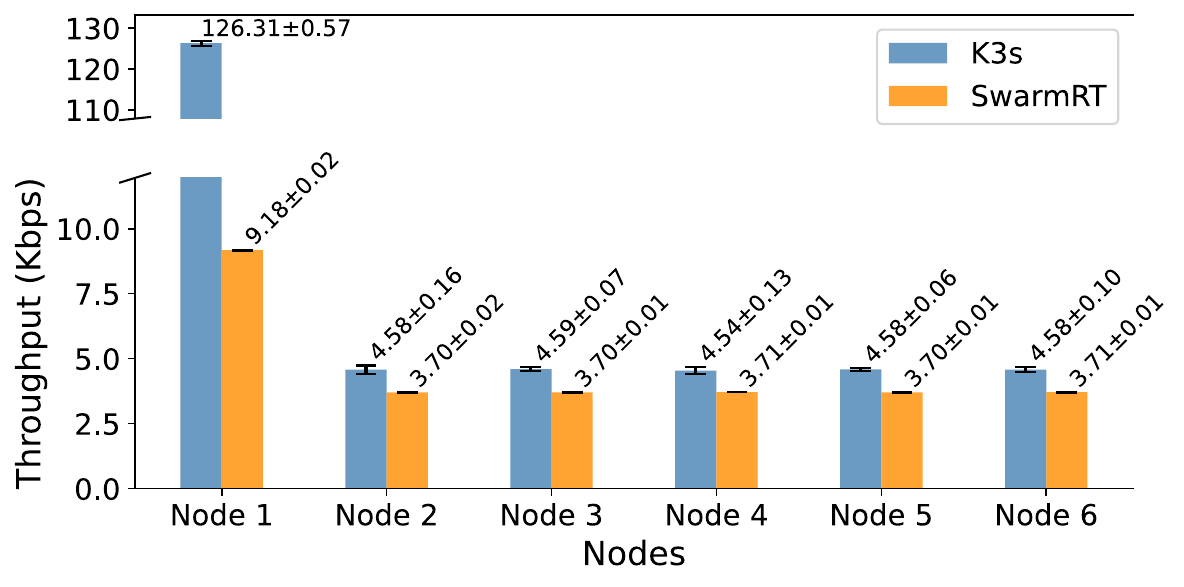}
        \caption{Throughput distribution over nodes}
        \label{fig:todo_traffic}
    \end{subfigure}

    \caption{Comparison of packet and traffic loads}
    \label{fig:network_throughput}
\end{figure}

Fig.~\ref{fig:todo_packet} compares the packet and traffic load of the two system architectures. In this experiment, we compare the network overhead of K3s and ROSCell under cells of identical size, since network utilization is a key indicator of scalability for large interconnected systems, and lower traffic also reduces data-transfer costs when devices communicate over cellular or Internet links. We use the same K3s version as in the previous section, with the default Kubernetes Container Network Interface (CNI), Flannel VXLAN. Node 1 in the figure is the master/coordination node, while Nodes 2--6 are the worker/primary nodes. The numbers above the bars represent the mean and standard deviation of the results of 10 trials.

In ROSCell, network traffic arises primarily from the Fast RTPS DDS simple discovery of the ROS2 coordinator node, as well as the DDS discovery client-server communication between workers and the master. Discovery traffic internal to the two ROSCell Docker containers is excluded, as it remains local to the device.

Following the previous setup, we used six Raspberry Pi 5 devices, one acting as a master and the remaining five as workers. Since we aim to characterize steady-state network behavior rather than short-term fluctuations caused by state transitions (e.g., nodes joining or leaving), we allow each cell to stabilize for two hours before data collection. For each run, we gathered 600 seconds of per-node network statistics and repeated this process ten times. 
As shown in Fig.~\ref{fig:network_throughput}, the ROSCell master node causes a network load more than 92\% lower than the K3s master node, while its worker nodes consume at least 18\% less network bandwidth compared to K3s workers.

\subsection{Multi-Object Pose Estimation}

We selected a multi-object pose estimation task commonly existing in multi-robot scenarios as an example use case to demonstrate the applicability of ROSCell and to examine different deployment strategies on edge devices.
For implementation, we adopted an AprilTag 3-based method\cite{krogius2019flexible}, which and its previous versions have been widely applied in robotic pose estimation \cite{nissler2016evaluation, kalaitzakis2021fiducial, moyo2024enhancing}. To further highlight ROSCell’s seamless integration with existing ROS2 software, we deployed the most-starred ROS2 AprilTag implementation on GitHub\footnotemark as a ROSCell skill.
\footnotetext{\url{https://github.com/christianrauch/apriltag_ros}}
\begin{figure}[htbp]
    \centering

    \begin{subfigure}{0.32\columnwidth}
        \centering
        \includegraphics[width=\columnwidth]{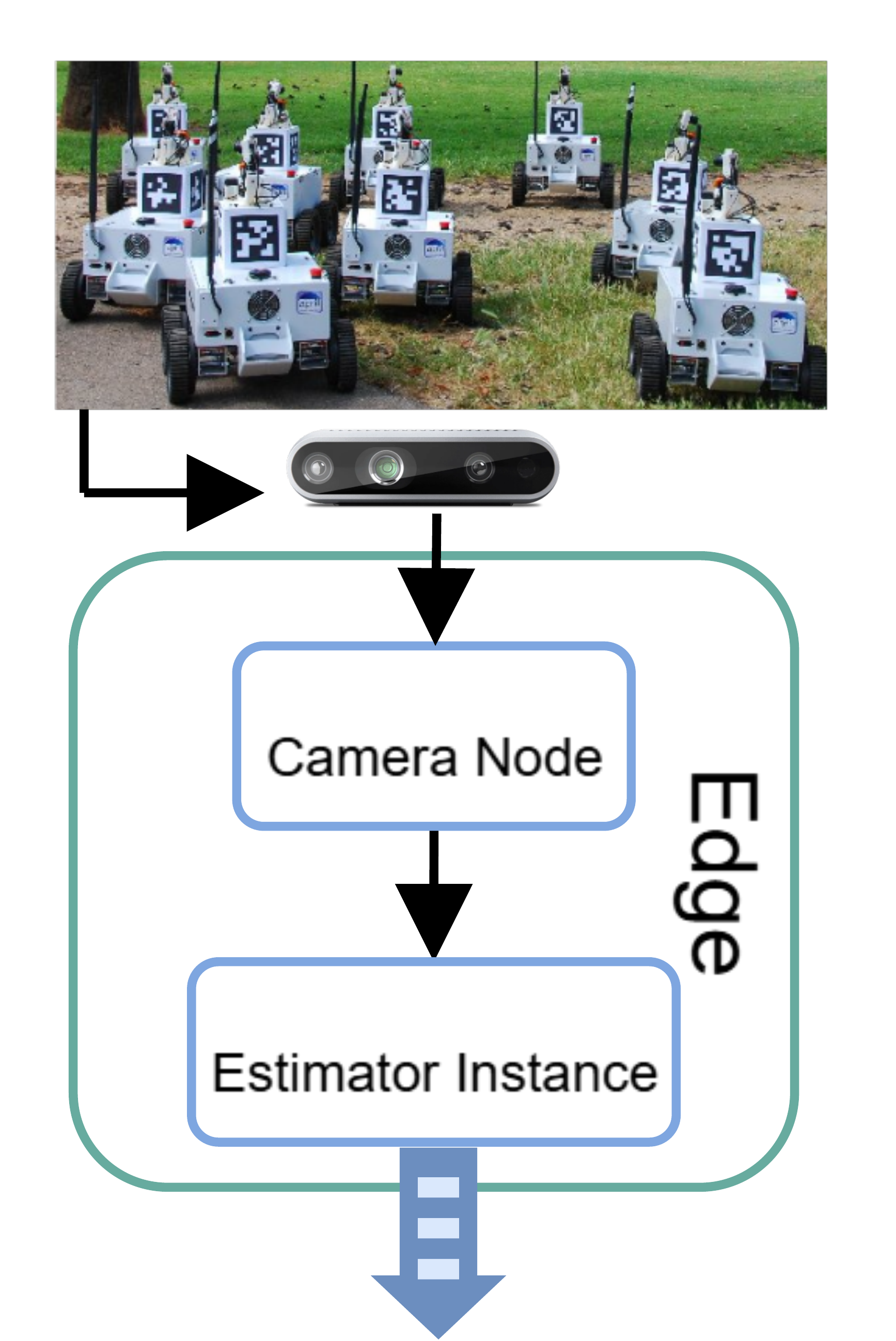}
        \caption{Use case example\protect\footnotemark\ }
        \label{fig:example_app}
    \end{subfigure}
    \hfill
    \begin{subfigure}{0.65\columnwidth}
        \centering
        \includegraphics[width=\columnwidth]{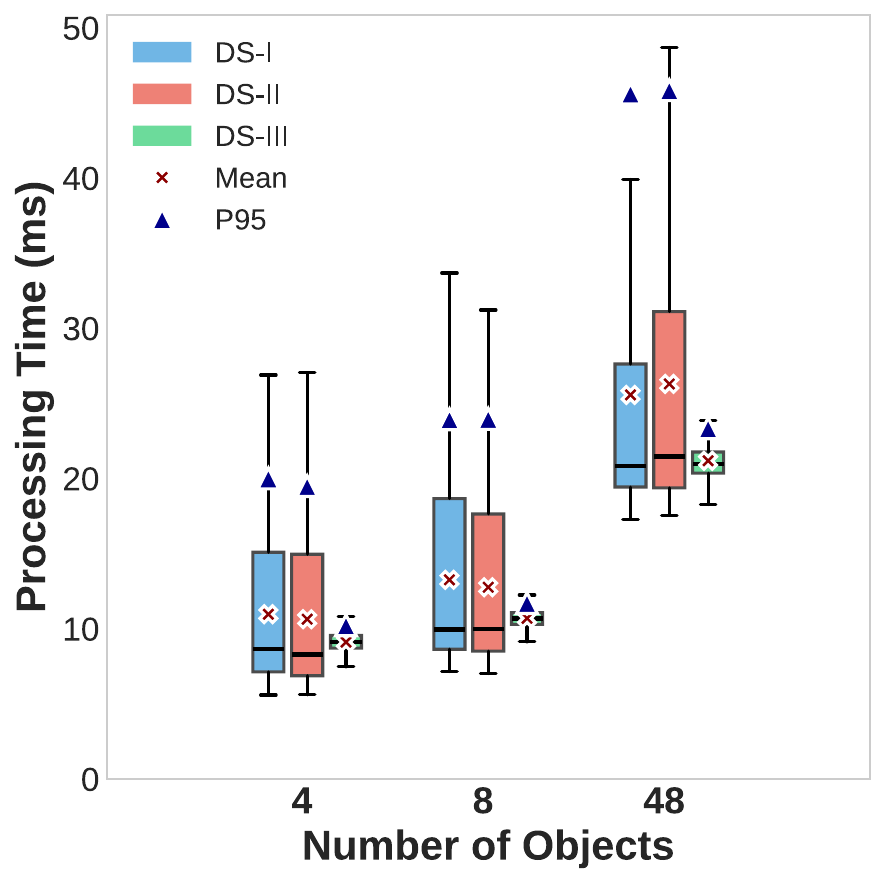}
        \caption{Comparison of deployment strategies}
        \label{fig:pose_estimation_res}
    \end{subfigure}
   
    \caption{\textbf{Multi-object pose estimation experiments:} a) illustrates an example use case in which robot poses are estimated from AprilTags by an edge device connected to the camera. The edge device functions as a ROSCell cell member, with the camera-stream ROS2 node and the pose-estimation skill instance running in two separate Docker containers. b) presents the per-frame processing time of the estimator under different deployment strategies.}
    \label{fig:multi_pose_est}
\end{figure}
\footnotetext{partially adapted from https://april.eecs.umich.edu/software/apriltag}
Fig. \ref{fig:example_app} presents a conceptual illustration of the example use case.
Our experiments used four scenes containing different numbers of AprilTags to represent robot fleets of varying sizes. The RealSense D435i camera streams image frames of each scene at 30 Hz.
We evaluated the per-frame processing time under three deployment strategies: DS-I) a single Raspberry Pi 5 running one instance to estimate poses for all four scenes; DS-II) a single Raspberry Pi 5 running four instances, each responsible for one scene; and DS-III) four Raspberry Pi 5 devices, each running one instance for a single scene. Each strategy was repeated ten times.
Notably, for fair comparison across strategies, each DS-III device is matched only with the processing unit in DS-I and DS-II that handles the same scene. For example, the scene-1 instance on device-1 in DS-III is compared with the scene-1 component in DS-I and the scene-1 instance in DS-II.

As shown in Fig. \ref{fig:pose_estimation_res}, DS-II outperforms DS-I when the number of scene objects is 4 or 8. This observation is consistent with findings in \cite{wen2024cloud}, which report that increasing the number of ROS nodes within a single container degrades ROS performance, whereas the isolation provided by Docker containers helps sustain more stable computational behavior.
However, when the number of objects increases to 48, the processing load becomes substantially heavier, and DS-I exhibits better stability than DS-II. This is because the limited computational capability of the edge device amplifies the overhead introduced by multiple containers, causing DS-II to experience more performance degradation under high-load conditions.
Across all scenarios, DS-III achieves the best performance, with stable processing-time distributions and p95 values consistently within the upper whiskers; its mean latency is also significantly lower than that of DS-I and DS-II. Although DS-III traditionally involves higher operational complexity due to multi-device deployment, this drawback is eliminated by ROSCell, which automates the entire deployment process with a single request.

\section{CONCLUSIONS}
We introduced ROSCell, a ROS2- and Docker-based framework for flexible formation and management of multi-robot systems. ROSCell abstracts heterogeneous devices into uniformly managed nodes and dynamically forms isolated cells as a computing continuum for automatic task deployment. It extends the K8s allocator and provides a task manager and skill library module, enabling seamless deployment of existing software across the cell.
Experiments show that ROSCell consumes significantly fewer resources—especially CPU and memory—than K8s/K3s-based frameworks, exhibits stronger role independence for heterogeneous device combinations, and generates substantially less network traffic. A multi-object pose estimation use case further demonstrates ROSCell’s ease of integrating ROS2 applications and executing complex deployment strategies.

Future work includes expanding ROSCell's capabilities, developing a user-friendly graphical interface, and exploring applications in robot learning, such as facilitating data privacy in continual learning \cite{lesort2020continual}.
In addition, aspects such as fault tolerance (e.g., automatic recovery from coordination node failures and skill instance crashes) and communication security (e.g., authentication and encryption across cells) are important for industrial deployment and will be addressed in future iterations of the framework.







\bibliographystyle{IEEEtran}
\bibliography{./references}

\end{document}